\documentclass[11pt]{article}

\usepackage[preprint]{acl}

\usepackage{times}
\usepackage{latexsym}

\usepackage[T1]{fontenc}

\usepackage[utf8]{inputenc}

\usepackage{microtype}

\usepackage{inconsolata}

\usepackage{graphicx}

\usepackage{amsmath}

\title{Calibration as a First-Class Criterion in LLM Evaluation}

\author{
    Mario Sanz-Guerrero$^1$ \and Katharina von der Wense$^{1,2}$ \\
    $^1$Johannes Gutenberg University Mainz, Germany \\
    $^2$University of Colorado Boulder, USA \\
    \texttt{\{\href{mailto:msanz@uni-mainz.de}{msanz}, \href{mailto:k.vonderwense@uni-mainz.de}{k.vonderwense}\}@uni-mainz.de}
}

\begin{document}
\maketitle
\begin{abstract}
Calibration of language models -- the alignment between expressed or implicit confidence and empirical correctness -- is a well-studied subfield within NLP.
Methods to measure it already exist.
The problem is adoption: outside this subfield, NLP research regularly introduces new models, datasets, and benchmarks without checking whether the model's confidence scores are meaningful. We argue that this adoption gap is a major obstacle to trustworthy LLM evaluation. Miscalibration causes problems in two distinct areas: at deployment, where overconfident mistakes cause real harm, and inside the research pipeline, where methods like LLM-as-a-judge, synthetic data generation, and active learning rely on calibrated confidence without verifying it. Standard calibration metrics 
only require two inputs per example: a confidence score and a correctness judgment. Most benchmarks in use today already provide both, meaning calibration can be reported immediately. For open-ended generation, however, defining these two inputs is still an open challenge. We argue that each NLP subfield should pair its main performance metric with a calibration score and call for treating calibration as an essential property of every model rather than a niche topic.
\end{abstract}

\section{Introduction} \label{sec:intro}

Large language models (LLMs) have moved from research prototypes to tools used by millions of people every day, and this shift changes how we need to evaluate them. Earlier NLP systems were narrow, task-specific models whose outputs were 
typically evaluated against a ground truth generated by domain experts.
In contrast, modern LLMs are general-purpose tools
that
can be applied to any task that takes text as input and produces text as output. Because of this versatility, millions of users now ask LLMs questions 
about a large variety of topics, and the model's confidence is the only indicator of the answer's expected correctness.
Further, outputs are increasingly not read by humans at all, but fed directly into autonomous agents that act on them without supervision. A benchmark score is therefore no longer just the end of an experiment -- it is the beginning of real-world deployment, 
where outputs may have severe consequences.

Performance metrics, such as accuracy or F1, answer a single question: did the model produce the correct output? They do not answer the practical question that deployment requires: should we trust this output? Two models that are right 90\% of the time are not interchangeable. A model whose confidence tracks its actual correctness is much more useful, because its 10\% errors are flagged rather than looking identical to its 90\% correct answers.

The property that separates these two models is calibration -- how well a model's stated or implicit confidence matches whether it is actually correct. Calibration is not a new idea. It has been studied for decades in statistics and classification \citep{brier1950, guo2017calibration}, and many recent NLP papers study it in LLMs \citep[\textit{inter alia}]{desai2020calibration,jiang2021know,kadavath2022languagemodelsmostlyknow,lin2022uncertaintywords,mielke2022overconfidence,tian2023verbalized,ulmer2024apricot,ulmer2026anthropomimetic}. A recent survey \citep{geng2024survey} organizes this literature. The methods are not the problem.

The problem is adoption. Outside the calibration subfield, NLP research often introduces new models, datasets, and benchmarks without measuring whether model confidence is meaningful. A machine translation paper reports BLEU \citep{papineni2002bleu}. A summarization paper reports ROUGE \citep{lin2004rouge}. An information extraction paper reports F1. A new benchmark publishes a leaderboard ranked only by accuracy. In each case, the question \emph{does the model know when it is wrong?} remains unanswered.

This gap is clear at the highest level of model development. We reviewed the public technical reports and model cards for recent releases across major model families: GPT-5.5 \citep{openai2026gpt5-5}, Claude Sonnet 4.6 \citep{anthropic2026sonnet4-6}, Gemini 3.5 Flash \citep{google2026gemini3-5}, DeepSeek V3.2 \citep{deepseekai2025deepseekv32}, Llama 3 \citep{grattafiori2024llama3}, Qwen3 \citep{yang2025qwen3}, Gemma 3 \citep{gemmateam2025gemma3}, GPT-OSS \citep{openai2025gptoss}, and OLMo 3 \citep{olmo2026olmo3}. All of them report results on dozens of capability and safety benchmarks, but \emph{none} reports calibration.\footnote{This review is meant as an illustration, not as a complete survey. We may have missed isolated cases, but the pattern across widely used models is consistent.} The GPT-4 technical report \citep{openai2024gpt4} is a notable earlier exception that documents how reinforcement learning from human feedback (RLHF) affects calibration, but later releases did not continue this practice.

We argue that this adoption gap is a major obstacle to trustworthy LLM evaluation. Calibration is not a specialized topic for a single subfield; it is a basic property of every model and should be evaluated as such. After defining calibration for LLMs (\S\ref{sec:definition}), we discuss three main points:
1)~Miscalibration causes problems in two places: at deployment, where overconfident errors cause concrete harm, and inside the research pipeline, where common practices (such as LLM-as-a-judge, synthetic data generation, and active learning) assume model confidence is calibrated without checking it (\S\ref{sec:stakes}).
2)~Existing calibration metrics need only two inputs per example: a confidence score and a correctness judgment. Most current benchmarks already provide both. Where metrics do not apply directly (such as open-ended generation), the challenge is defining these two inputs, not creating entirely new metrics (\S\ref{sec:metrics}).
3)~Closing this gap requires two steps that can happen in parallel: adopting community reporting standards for the tasks where metrics already work today and researching how to define confidence and correctness for open-ended generation (\S\ref{sec:directions}).

\section{Calibration for LLMs}
\label{sec:definition}

A predictor is calibrated if, among the predictions it makes with confidence $p$, a fraction $p$ are correct \citep{guo2017calibration}. This is a population-level property and is separate from accuracy: a model that always predicts with confidence 0.7 and is correct 70\% of the time is perfectly calibrated, even though we do not know in advance which individual answers are right. Measuring calibration requires two pieces of information for each example: a confidence score and a judgment of whether the output is correct. The correctness judgment usually comes directly from the task. The confidence score is less straightforward, because LLMs express confidence in at least three ways.

\paragraph{Token and sequence probabilities.} An autoregressive LLM defines a distribution over output sequences as
\[
p(y \mid x) = \prod_{t=1}^{T} p(y_t \mid x, y_{<t}),
\]
where each factor is the probability the model assigns to token $y_t$ at step $t$.
The sequence probability, often length-normalized as $p(y \mid x)^{1/T}$ to compare outputs of different lengths, is the natural extension of classifier confidence to generation, and it is where early calibration studies of transformer-based models began \citep{desai2020calibration, jiang2021know}. It is also the only confidence signal that exists by construction; verbalized confidence and behavioral cues must be elicited or interpreted. Even in simple multiple-choice QA, extracting this signal involves design choices (e.g., which tokens represent the answer or how to account for answer length) that affect the measured confidence \citep{sanzguerrero2025mind, sanzguerrero2025lengthbias}.

\paragraph{Verbalized confidence.} The model is prompted to state its confidence in words (e.g., ``I am 80\% sure''). Recent work shows that this signal can be elicited for any task and that for instruction-tuned models it is sometimes better calibrated than raw probabilities \citep{lin2022uncertaintywords, tian2023verbalized}.

\paragraph{Behavioral signals.} Models also indicate confidence through behavior, such as hedging, refusing to answer, or expressing doubt. These are implicit signals that users actually read and interpret.

~\\
Token and sequence probabilities, verbalized confidence, and behavioral signals are not interchangeable. A model can have well-calibrated token probabilities but poorly calibrated verbalized confidence, or the reverse \citep{kadavath2022languagemodelsmostlyknow, tian2023verbalized}. In deployment, users and downstream components (such as autonomous agents) only see the generated text, not the internal softmax probabilities. Verbalized and behavioral calibration are therefore what users actually rely on, while token-level calibration remains important for model analysis, training, and selective-prediction systems that have direct access to log-probabilities. Any evaluation of calibration should clearly state which of these signals is being tested.

A second important distinction \citep{kendall-gal-2017} separates the \emph{sources} of uncertainty. \emph{Aleatoric} uncertainty is irreducible: it comes from ambiguity in the input itself, such as a question with multiple valid answers or under-specified context. \emph{Epistemic} uncertainty is reducible: it reflects the model's lack of knowledge, which could shrink with more training data or better retrieval. Standard calibration metrics treat both types the same, but the distinction matters in practice because each calls for a different response -- abstention for aleatoric uncertainty and retrieval or further training for epistemic uncertainty.

\section{Why Calibration Failures Matter}
\label{sec:stakes}

Below, we discuss three settings where poor calibration causes problems, followed by an explanation of why miscalibration continues to grow.

\paragraph{Human--AI interaction in high-stakes domains.} Users naturally adjust their trust based on how confident a model appears \citep{steyvers2025know}. People are likely to act on a wrong answer if it sounds confident, but will double-check a correct answer if the model sounds hesitant \citep{kim2024notsure,zhou2024relying}. 
For example, in legal applications, evaluations of leading LLMs show hallucinated case citations and fabricated court decisions delivered with complete confidence \citep{dahl2024legal}.
In medical question answering, hallucinated clinical facts and incorrect drug dosages remain a frequent failure mode \citep{kim2025medical}, and non-expert users cannot easily detect them. In both settings, the real danger is not just that the model makes mistakes, but that it gives no warning when it does. A wrong answer is far more dangerous when expressed with absolute certainty than when presented with appropriate doubt. The opposite behavior is not helpful either: a model that hedges on every single response provides no useful signal. Both cases are calibration failures.

\paragraph{Agentic and reasoning systems.} When LLMs are chained together in agentic pipelines (e.g., planner, retriever, and executor), confidence is the signal that tells the system whether to take an action, ask for user input, or stop. If one component is miscalibrated, its overconfident mistakes propagate directly into subsequent steps \citep{el-yaniv-wiener-2010}.
In automated systems, frontier models can take actions at very low probabilities \citep{serrano2026frontiermodelsactionslow}, and standard calibration metrics will not see them.
A similar problem happens inside reasoning models that generate step-by-step chains of thought. Mistakes build on each other: an overconfident error early in a reasoning trace often leads to an incorrect final response. Measuring calibration only on the final answer misses these internal mistakes entirely. Evaluation should therefore examine the calibration of the entire reasoning trace, not just the final output \citep{yoon2025reasoning}.

\paragraph{The research pipeline.} Miscalibration does not just cause problems during deployment; it also damages the research process itself. Several common practices in NLP assume that model confidence is meaningful and fail when it is not. First, \emph{LLM-as-a-judge} evaluation uses one model to score the outputs of another. If the judge is miscalibrated, the resulting rankings, win rates, and reported improvements are biased. Second, \emph{synthetic data generation} uses LLMs to create new training corpora. A miscalibrated generator produces confident errors that the next round of training then learns from. Finally, \emph{active learning, data filtering, and uncertainty-guided retrieval} all select examples based on confidence scores, so miscalibrated confidence means selecting the wrong data points. 

\subsection{Increased Miscalibration: Post-Training Degrades Calibration}
Training optimizes what we measure, and we (generally) do not measure calibration. Base models are reasonably well calibrated on multiple-choice tasks. However, instruction tuning and RLHF hurt calibration, even when accuracy improves \citep{openai2024gpt4}. Part of this issue comes from the conversational format itself: instruction-tuned models are significantly more confident in an answer when it is presented to them as their own output than when the same answer is provided by the user \citep{sanzguerrero2026overconfident}. RLHF can also lead to increased rates of sycophancy \citep{sharma2024sycophancy}, where models adjust their confidence to agree with the user's beliefs instead of reflecting whether they are actually right. Furthermore, \citet{kalai2025languagemodelshallucinate} point out that most benchmarks give the same zero score to saying ``I don't know'' as they do to an incorrect answer. As a result, guessing blindly is strictly preferable to abstaining, so current training and evaluation setups reward confident guessing, which directly promotes hallucinations.
When we optimize solely for headline accuracy, we end up damaging calibration because it remains unmeasured. Until we treat calibration as a first-class evaluation criterion, standard training pipelines will continue to degrade it.

\section{Current Metrics and Their Limits}
\label{sec:metrics}

Below, we summarize standard calibration metrics and explain where each falls short for LLMs. All of these metrics require the same two inputs per example: a confidence score $\hat{p}_i$ and a correctness label $y_i$. The mathematical formulation of the metrics does not depend on whether the task is classification or generation. What changes across tasks is how easy or difficult it is to define these two inputs.

\paragraph{Expected Calibration Error.}
ECE partitions predictions into $M$ confidence bins and reports the weighted average gap between bin accuracy and bin confidence:
\[
\mathrm{ECE} = \sum_{m=1}^M \frac{|B_m|}{N} \bigl|\, \mathrm{acc}(B_m) - \mathrm{conf}(B_m) \,\bigr|,
\]
where $B_m$ is the set of predictions in bin $m$ and $N$ is the total number of predictions \citep{naeini2015ece, guo2017calibration}.
The same bins give the \emph{reliability diagram}, which plots bin accuracy against bin confidence: a perfectly calibrated model lies on the diagonal, points below it indicate overconfidence, and points above it indicate underconfidence.
Two limitations are especially important for LLMs. First, ECE estimates are bin-sensitive and statistically biased \citep{kumar2019verified}, and the reported value depends on binning choices that are rarely justified.
Second, ECE assumes a single numerical confidence score for each prediction over a fixed set of classes. For open-ended generation, defining ``the prediction'' and ``its confidence'' is not straightforward.

\paragraph{Brier score.} For a binary outcome $y \in \{0,1\}$ with predicted probability $\hat{p}$, the Brier score \citep{brier1950} is the mean squared error over $N$ predictions:
\vspace{-0.5\baselineskip}
\[
\mathrm{BS} = \frac{1}{N}\sum_{i=1}^N (\hat{p}_i - y_i)^2
\]
Unlike ECE, which can be pushed toward zero simply by predicting the overall dataset accuracy, the Brier score is a \emph{proper scoring rule}: it is minimized only when the predicted probabilities match the true empirical frequencies. However, like ECE, the Brier score assumes discrete outcomes. Applying it to free-form text requires simplifying each generated response into a binary correct-or-incorrect label, which leaves out important nuances in open-ended answers.

\paragraph{AUROC and selective prediction.} AUROC measures the probability that a randomly selected correct prediction receives a higher confidence score than a randomly selected incorrect one: 
\[
\mathrm{AUROC} = \Pr\bigl(\hat{s}(x^+) > \hat{s}(x^-)\bigr),
\]
where $\hat{s}$ is the confidence score, $x^+$ is a correctly classified input, and $x^-$ is an incorrectly classified one. Related evaluation curves, e.g., the accuracy--rejection curve, measure how much accuracy improves when the model abstains from answering low-confidence predictions \citep{el-yaniv-wiener-2010}. Because AUROC depends only on the ranking of confidence scores rather than their numerical values, a model that inflates all its confidences by the same amount keeps the same AUROC. Thus, AUROC measures ranking (how well confidence separates correct from incorrect answers) rather than calibration (whether the confidence numbers themselves are meaningful), which is less interpretable and less useful for deployment.

\paragraph{Where the metrics apply.} For confidence, verbalized estimates can be elicited on almost any task and scored with the metrics above \citep{lin2022uncertaintywords, tian2023verbalized, xiong2024uncertainty}, although they are sensitive to prompt phrasing, lack standardization across benchmarks, and mix two questions: whether the model internally knows it is uncertain, and whether it expresses that uncertainty accurately in words. Sequence probabilities are available whenever there is a single canonical target. For correctness, subfields already rely on established criteria: exact match in question answering, unit test pass rates in coding, or verified final answers in mathematics. Whenever this correctness check is binary (or can be made binary using a standard threshold), existing calibration metrics work directly. This applies to most benchmarks featured in the technical reports from Section~\ref{sec:intro}, which focus on multiple-choice, short-answer, math, and code generation tasks. Where standard metrics fail is open-ended generation: when many different answers are valid, there is no single target sequence whose probability we can measure, making both confidence and correctness harder to define. We turn to this open problem next.

\section{Directions}
\label{sec:directions}

Below, we separate what can be done now from what still needs research, and close with one direction beyond calibration.

\paragraph{Calibration in every subfield.} Every subfield in NLP has its own standard metrics: BLEU and COMET in machine translation, ROUGE in summarization, F1 in information extraction, win rates in instruction following, and accuracy in QA. Each task should pair its primary metric with a calibration score that measures whether model confidence actually tracks performance. Doing this simply requires choosing a reasonable confidence signal, reusing the correctness criteria the subfield already relies on, and adding a column to the results table. Machine translation already shows this is possible: quality estimation predicts translation quality without a reference \citep{specia2018qe}, serving as an effective confidence signal. Yet quality estimation scores are rarely reported alongside BLEU as an intrinsic property of the translation model.
The reason this is not standard practice is convention, not difficulty. The same convention explains why recent model releases (discussed in Section~\ref{sec:intro}) report scores across dozens of capability benchmarks, but leave out calibration entirely.

\paragraph{Reporting norms.} Community standards are the best way to drive change, and we propose two concrete changes. First, every benchmark result should include a calibration score alongside its main score, and major leaderboards should add a column for it. Second, reviewers should treat the absence of such reporting as a methodological gap, comparable to leaving out basic training settings. \citet{kalai2025languagemodelshallucinate} suggest a related idea: change how benchmarks are scored so that being confidently wrong hurts a model's score more than saying ``I don't know.'' Instead of adding a new column, this changes what current leaderboards measure, and both ideas are compatible.

\paragraph{Calibration for free-form generation.} Most modern LLM applications involve open-ended generation, where confidence and correctness are not yet clearly defined, and this is the area that still needs research.
Grouping generated responses by meaning rather than surface wording \citep{kuhn2023semantic} is a starting point. However, standardizing approaches and analyzing how they perform across tasks remains open. Crucially, this research should move forward in parallel with reporting norms on simpler tasks, rather than delaying them.

\paragraph{Verbalized confidence as an evaluation target.} In practical applications, users interact directly with a model's generated text, including any stated confidence or doubts. The NLP community should therefore treat verbalized confidence as an evaluation target in its own right, developing standardized prompt formats, consistent scoring methods, and analyses that distinguish between what a model internally knows and what it actually says. \citet{ulmer2026anthropomimetic} take this a step further, arguing that verbalized uncertainty should reflect natural human communication. Because users interpret model statements the same way they interpret human conversation, a model whose numerical probabilities are accurate but whose expression of uncertainty sounds unnatural might still mislead readers.

\paragraph{Beyond calibration: attribution.} Calibration answers one fundamental question about trust: \emph{when} should we believe an output? A second question is \emph{why}: what evidence supports it? For LLMs, training data provides this evidence, and attribution methods aim to identify which training examples most influenced a particular output \citep{koh2017influence, grosse2023influence}. Calibration and attribution complement each other well: calibration gives users a score indicating how much to trust an answer, while attribution provides verifiable evidence showing whether a confident response is genuinely grounded in training data. While attribution methods are not yet as mature as calibration metrics, they face the exact same adoption challenge. Once these methods become practical, attribution should also be reported as a standard property of every model rather than treated as a niche experiment.

\section{Conclusion}

As LLMs move from research to widespread use, our evaluation practices must change as well. 
Practical methods to evaluate calibration already exist, but mainstream NLP research often overlooks them. This adoption gap creates risks in deployment when overconfident mistakes go unnoticed, and it undermines research pipelines that rely on uncalibrated models. For most benchmarks, the necessary calibration metrics are already available -- all that is missing is the community standard to report them. For open-ended generation, defining appropriate metrics remains an important research challenge.

\section*{Limitations}

This is a position paper without new empirical experiments; our arguments build on findings from existing literature. In addition, calibration for open-ended generation does not yet have a consensus definition. We focus our concrete reporting proposals on tasks where calibration metrics are already well established, while emphasizing that developing metrics for free-form generation remains an essential area for future research.

\section*{Acknowledgments}
This work was supported by the Carl Zeiss Foundation through the MAINCE project (grant number P2022-08-009).

\bibliography{refs}

\end{document}